\documentclass{article}
\pdfoutput=1

\usepackage[final,nonatbib]{neurips_2022}

\usepackage[utf8]{inputenc} % allow utf-8 input
\usepackage[T1]{fontenc}    % use 8-bit T1 fonts
\usepackage{url}            % simple URL typesetting
\usepackage{booktabs}       % professional-quality tables
\usepackage{amsfonts}       % blackboard math symbols
\usepackage{nicefrac}       % compact symbols for 1/2, etc.
\usepackage{microtype}      % microtypography
\usepackage{xcolor}         % colors
\usepackage{colortbl}
\usepackage{xspace}
\usepackage{comment}
\usepackage{float}

\usepackage{caption}
\usepackage{subcaption}
\usepackage{graphicx}

\usepackage[symbol]{footmisc}
\renewcommand{\thefootnote}{\fnsymbol{footnote}}

\usepackage{hyperref}
\newcommand{\OM}[0]{\texttt{Comp2Comp}\xspace}

\usepackage{pmboxdraw}
\usepackage{etoolbox}
\makeatletter
\AtBeginEnvironment{minted}{\dontdofcolorbox}
\def\dontdofcolorbox{\renewcommand\fcolorbox[4][]{##4}}
\makeatother

\usepackage{xcolor}
\hypersetup{
    colorlinks,
    citecolor=gray,
    linkcolor=black,
    urlcolor=black
}

\definecolor{cosmiclatte}{rgb}{1.0, 0.97, 0.91}

\usepackage[frozencache=true,cachedir=minted-cache]{minted}
\newenvironment{codeshell}
 {\VerbatimEnvironment
 \begin{minted}[fontsize=\footnotesize,bgcolor=cosmiclatte]{shell}}
 {\end{minted}}

\title{Comp2Comp: Open-Source Body Composition Assessment on Computed Tomography}

\author{
Louis Blankemeier$^{*1}$ \quad Arjun Desai$^{*1}$ \quad Juan Manuel Zambrano Chaves$^{1}$ \quad \\
\textbf{Andrew Wentland}$^{2}$ \quad \textbf{Sally Yao$^{1}$} \quad \textbf{Eduardo Reis$^{1}$} \quad \textbf{Malte Jensen$^{1}$} \\
\quad \textbf{Bhanushree Bahl$^{3}$} \quad \textbf{Khushboo Arora$^{3}$} \quad \textbf{Bhavik N. Patel$^4$} \\ \quad \textbf{Leon Lenchik$^{5}$} \quad \textbf{Marc Willis$^{1}$} \quad
\textbf{Robert D. Boutin$^{1}$} \quad  \textbf{Akshay S. Chaudhari$^{1}$}
\vspace{3mm}
\\
$^1$Stanford University \quad $^2$University of Wisconsin-Madison \\
$^3$CARPL.ai \quad $^4$Mayo Clinic \quad $^5$Wake Forest University \vspace{3mm}
\\
\texttt{\{lblankem,arjundd,jmz,yaohanqi,edreis,mekj,} \\
\texttt{marc.willis,boutin,akshaysc\}@stanford.edu} \\
\texttt{alwentland@wisc.edu} \\
\texttt{\{bhanushree.bahl,khushboo.arora\}@carpl.ai} \\
\texttt{patel.bhavik@mayo.edu} \\
\texttt{llenchik@wakehealth.edu} 
}

\begin{document}

\maketitle

\def\thefootnote{*}
\footnotetext{These authors contributed equally to this work}
\newcommand\blfootnote[1]{
  \begingroup
  \renewcommand\thefootnote{}\footnote{#1}
  \addtocounter{footnote}{-1}
  \endgroup
}

\begin{abstract}
Computed tomography (CT) is routinely used in clinical practice to evaluate a wide variety of medical conditions. While CT scans provide diagnoses, they also offer the ability to extract quantitative body composition metrics to analyze tissue volume and quality. Extracting quantitative body composition measures manually from CT scans is a cumbersome and time-consuming task. Proprietary software has been developed recently to automate this process, but the closed-source nature impedes widespread use. There is a growing need for fully automated body composition software that is more accessible and easier to use, especially for clinicians and researchers who are not experts in medical image processing. To this end, we have built \href{https://github.com/StanfordMIMI/Comp2Comp}{\texttt{Comp2Comp}}, an open-source Python package for rapid and automated body composition analysis of CT scans. This package offers models, post-processing heuristics, body composition metrics, automated batching, and polychromatic visualizations. \OM currently computes body composition measures for bone, skeletal muscle, visceral adipose tissue, and subcutaneous adipose tissue on CT scans of the abdomen. We have created two pipelines for this purpose. The first pipeline computes vertebral measures, as well as muscle and adipose tissue measures, at the T12 - L5 vertebral levels from abdominal CT scans. The second pipeline computes muscle and adipose tissue measures on user-specified 2D axial slices. In this guide, we discuss the architecture of the \OM pipelines, provide usage instructions, and report internal and external validation results to measure the quality of segmentations and body composition measures. \OM can be found at \url{https://github.com/StanfordMIMI/Comp2Comp}.
\end{abstract}

%, including those related to cancer, trauma, and infection.

\section{Motivation}
Computed tomography (CT) is a widely used volumetric medical imaging modality, with approximately 80 million CT scans performed annually in the US, of which the most frequently imaged anatomy is the abdomen~\cite{radiol,bellolio2017increased,hess2014trends}. As CT continues to be an important tool for qualitatively evaluating patient health and detecting disease, the number of CT scans performed is expected to increase~\cite{doi:10.7326/0003-4819-158-8-201304160-00003,liu2022fully}. While CT scans provide qualitative diagnostic insights, they also offer the ability to extract quantitative body composition metrics of tissue quantity and quality. These quantitative metrics can provide diagnostic and prognostic biomarkers for both acute and chronic health conditions~\cite{lee2022abdominal,thibault2012body,kuriyan2018body}, as well as be followed longitudinally to evaluate for positive or negative trends. By accurately segmenting bone, muscle, and adipose tissue, CT enables more accurate evaluation of body composition than traditional clinical measurements (e.g., weight, body mass index (BMI), waist circumference, skinfolds)~\cite{zeng2021ct}. However, manually extracting objective, quantitative body composition measures from CT scans is a cumbersome task taking several minutes which is not practical during the clinical practice of medicine or for large-scale research studies.

We introduce \OM, which is an open-source Python package designed to simplify and expedite CT-based body composition analysis. \OM refers to the transformation of routine “computed tomography to body composition” data. Specifically, this package contains methods to automatically segment CT images, compute and manage CT-based body composition measures, and visually display polychromatic output for quality assurance purposes. Our package operates directly on the DICOM medical image standard and produces body composition outputs in several formats, including a visual report. We provide two pipelines for different use cases. The first pipeline computes 3D spine metrics and multi-level muscle and adipose tissue metrics at T12 - L5. The second pipeline forgoes spine analysis and performs muscle and adipose tissue analysis on user-specified 2D axial slices. The package is designed to be easy to install and provides an intuitive command line interface. The package is hosted on the GitHub platform at https://github.com/StanfordMIMI/Comp2Comp and is freely available under the AGPL-3.0 License.

\OM currently computes body composition measures on routine abdominal CT scans for vertebral trabecular bone, total abdominal skeletal muscle, visceral adipose tissue (VAT), and subcutaneous adipose tissue (SAT) on CT scans of the abdomen. 

Bone evaluation is important because it allows screening for osteoporosis (low bone mass and quality that results in increased fracture risk). Osteoporosis is underdiagnosed and undertreated, even though effective treatments are available and  complications (e.g., major fractures, premature death) are widely recognized.  Currently, there is a profound unmet clinical need for implementing improved screening to facilitate appropriate treatment~\cite{boutin2020value}. Previous retrospective studies have established that trabecular bone density of $<90$ Hounsfield units (HU) at the L1 vertebral body is associated with a high risk of vertebral fracture (odds ratio, 32) and predicts future fractures throughout the body~\cite{graffy2017prevalence,lee2018future}. Large scale screening for osteoporosis using CT has been validated in retrospective cohorts~\cite{roux2022fully,pickhardt2020automated}, but is not yet widely implemented because automated techniques have not been freely disseminated. Our automated spine segmentation and bone density analysis package is open source, and should facilitate wide scale adoption. 

Skeletal muscle evaluation can also add value to routine abdominal CT scans by screening for sarcopenia. Sarcopenia, broadly defined by the loss of normal muscle tissue and muscle function, is associated with numerous adverse clinical outcomes, including frailty, post-operative complications, and premature death. Sarcopenia prevalence increases with age and many diseases (ranging from approximately 10\% prevalence in community-dwelling adults over the age of 60 years to 24\% in hospitalized patients to 35\% in cancer patients)~\cite{papadopoulou2020differences,surov2022prevalence}. On routine CT, sarcopenia can manifest as low muscle volume (myopenia) or low muscle density (myosteatosis). Manual segmentation of muscle on CT is impractical for real-time clinical implementation, but automated algorithms would allow for treatment interventions earlier in the disease course while a favorable window of anabolic potential is open~\cite{nowak2021end}.

Comp2Comp also analyzes VAT and SAT on routine abdominal CT scans. Similar to muscle, adipose tissue depots can be segmented to enable measurements of tissue quantity (e.g., cross-sectional area measured in cm$^2$) and quality (e.g., density measured in HU). Adipose tissue, particularly VAT, is a metabolically active tissue that is a modifiable risk factor for numerous medical conditions, including metabolic syndrome, nonalcoholic fatty liver disease~\cite{vilalta2022adipose}, heart failure~\cite{rao2021regional}, future cardiovascular events~\cite{chaves2021opportunistic}, cancer~\cite{katzmarzyk2022association}, and cancer mortality~\cite{katzmarzyk2022association}. Furthermore, VAT density on CT scans is associated with all-cause mortality, as well as death related to cancer and cardiovascular disease~\cite{rosenquist2015fat}. As with other tissues evaluated on routine abdominal CT scans, there are increasingly sophisticated reference values that account for demographics (e.g., sex, ethnicity) that can be used to define specific cut-off values for abnormalities in muscle~\cite{derstine2018skeletal,derstine2018quantifying} and VAT~\cite{baggerman2022computed,derstine2022healthy}.

\section{Installation}
We provide an installation script that sets up an Anaconda / Miniconda environment with all necessary dependencies. As such, we assume that Anaconda or Miniconda is installed. We recommend that users use this method to install \OM. To install \OM, run the following:

\begin{codeshell}
git clone https://github.com/StanfordMIMI/Comp2Comp/
cd Comp2Comp && bin/install.sh
\end{codeshell}

Alternatively, the package can be installed with pip using the command below.

\begin{codeshell}
git clone https://github.com/StanfordMIMI/Comp2Comp/
cd Comp2Comp && pip install -e .
\end{codeshell}

\section{Pipeline 1: End-to-End Spine, Muscle, and Adipose Tissue Analysis at T12 - L5}
\begin{figure}[ht]
    \centering
    \includegraphics[scale=0.28]{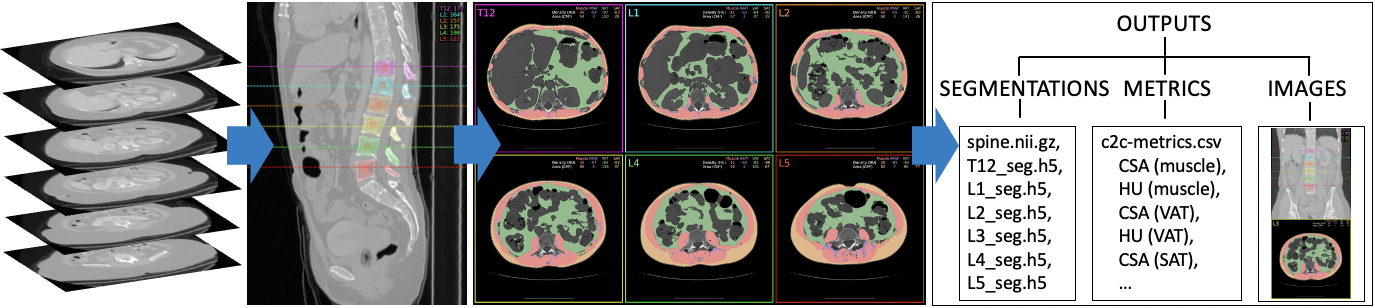}
    \caption{The path to a folder containing a DICOM series or subfolders with DICOM series is provided as input. 3D spine segmentation of the T12 - L5 vertebral levels is then performed separately on each of these DICOM series. Using the segmentation predictions, we identify 3D regions of interest (ROIs) from which to extract a measure of Hounsfield units (HUs) from the trabecular bone of the vertebral bodies. The HU measures from these ROIs provide surrogates of bone mineral density (BMD). Additionally, using the spine segmentations, we determine superior-inferior centers of each vertebral level, represented by the dashed lines the sagittal image. We extract the DICOMs corresponding to superior-inferior center and then perform 2D skeletal muscle and adipose tissue segmentation at each of these levels. Finally, we save outputs that include segmentation files, metrics, and polychromatic images.}
    \label{fig:workflow}
\end{figure}

\OM enables quantification of bone, muscle, and adipose tissues on abdominal CT scans between the T12 and L5 vertebral levels. We depict the workflow of \emph{Pipeline 1} in Figure~\ref{fig:workflow}. 

\subsection{Inputs}
The information generated by CT scans are stored as digital imaging and communications in medicine (DICOM) files~\cite{parisot_1995}, where each file is composed of a header and image data. The header contains metadata that are used to describe equipment settings and information about the patient, while the image data stores the CT pixel information. To facilitate integration with the current clinical workflow, DICOM images serve as the input to our pipeline. In particular, for \emph{Pipeline 1}, the inputs are volumetric CT data stored as series of DICOM files.

\subsection{Spine Segmentation}
\noindent\textbf{Model:} In our initial release (v0.1.0), we use the spine-only model from TotalSegmentator~\cite{https://doi.org/10.48550/arxiv.2208.05868} for spine segmentation. This is a 3D nnU-Net model~\cite{isensee2021nnu} trained on a dataset of 1,082 CT scans with the C1 - L5 vertebrae labeled (for Comp2Comp, we use the T12 - L5 segmentations). For more details about this method, model, and training strategy, refer to \cite{https://doi.org/10.48550/arxiv.2208.05868}.

\begin{figure}[ht]
    \centering
    \includegraphics[scale=0.43]{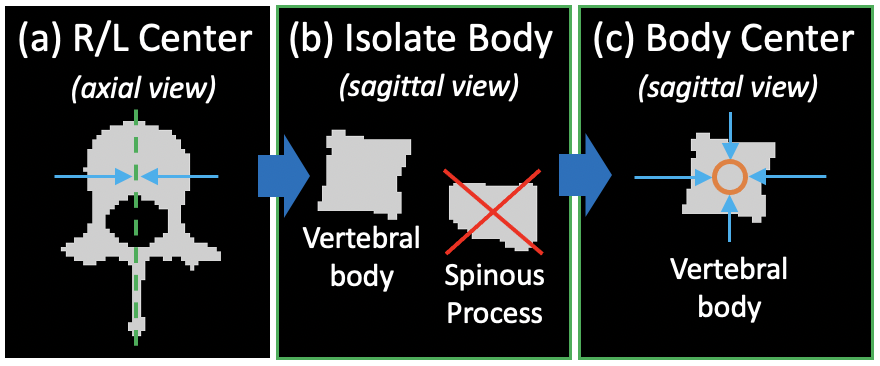}
    \caption{This figure demonstrates our process for generating spine ROIs. Note that the jagged edges are due to the resolution of the spine segmentation model. (a) shows how we compute the right-left center of mass for each 3D per-level segmentation label. For each vertebral level, we then select the two largest connected components in the sagittal planes through the right-left centers, extract the two largest 2D connected components within these planes, and isolate the 2D vertebral body segmentations as shown in (b). Finally, we compute the center of masses for each level within the 2D vertebral body segmentations. We construct a 3D ROI around the 3D centers generated by the vertebral body centers combined with the right-left centers as shown in (c).}
    \label{fig:spine_rois}
\end{figure}

\noindent\textbf{Spine Regions of Interest and Trabecular HU:} Trabecular bone regions of interest (ROIs) are computed at each T12 - L5 vertebral level. To identify the ROIs, we compute the right-left center of mass of the segmentation prediction for each vertebral level separately (panel (a) in Figure~\ref{fig:spine_rois}). The sagittal plane through this center should pass through the spinal canal, separating the vertebral body from the spinous process. On this slice, we take the two largest connected components which should correspond to the vertebral body and the spinous process. Based on the centroids of these two connected components, we discard the posterior-most connected component, corresponding to the spinous process (panel (b) in Figure~\ref{fig:spine_rois}). Then, for each vertebral level and corresponding sagittal slice, we compute a 2D center of mass of the vertebral body in the 2D sagittal plane (panel (c) in Figure~\ref{fig:spine_rois}). In our initial release (v0.1.0), we construct a 1 cm diameter sphere (optionally cubic) centered on the 3D center generated by combining the right-left center with the 2D vertebral body center. This enables us to compute an HU measure within these ROIs for each vertebral level. These HU measures can be computed using the mean or median of pixels within the ROIs. We hypothesize that the median provides a measure that is more robust to the unintentional presence of cortical bone in the ROIs.

\noindent\textbf{Superior-Inferior Centers:} For each vertebral level (T12 - L5), we compute the superior-inferior centers of the foreground pixels, including the spinous/transverse processes. These centers are then used to select 6 axial DICOM files, one per vertebral level, which are subsequently passed to the muscle and adipose tissue segmentation model for further processing.

\subsection{Muscle and Adipose Tissue Segmentation}
\noindent\textbf{Models:} We provide two models for 2D segmentation of muscle and adipose tissue.  Both models are two-dimensional UNet convolutional neural networks (CNNs). These models were trained on axial DICOM slices at the L3 vertebral level. The models are hosted on Hugging Face, and are publicly available at \url{https://huggingface.co/stanfordmimi/stanford_abct_v0.0.1/blob/main/stanford_v0.0.1.h5} and \url{https://huggingface.co/stanfordmimi/stanford_abct_v0.0.1/blob/main/abCT_v0.0.1.h5}. The \verb|stanford_v0.0.1| model segments muscle, cortical bone (not used by Comp2Comp), VAT, and SAT. The \verb|abCT_v0.0.1| model segments muscle, intermuscular adipose tissue (IMAT), VAT, and SAT. Further details about the training strategy are provided in~\cite{chaves2021opportunistic}.

\noindent\textbf{Post-Processing:} By default, we perform post-processing to fill holes in the segmentation masks that are less than 200 pixels in area for SAT and less than 20 pixels in area for other tissues. These numbers were chosen based on visual inspection and can be modified. We perform additional post-processing to compute body composition metrics for IMAT. This is done for both the \verb|stanford_v0.0.1| and \verb|abCT_v0.0.1| models. To do this, we change the label of a given pixel to IMAT if it was originally labeled as muscle and has an HU value that is $<-30$ and $>-190$. To reduce the impact of noisy pixels being falsely identified as IMAT pixels, we filter out any connected components with a size of less than 10 pixels. It is important to note that the \verb|abCT_v0.0.1| model also outputs IMAT predictions based on training with IMAT labels, making it better suited for IMAT analysis.

\noindent\textbf{Muscle and Adipose Tissue Metrics:} Using the post-processed segmentation predictions, we can identify various tissues within the images and calculate their estimated density using the Hounsfield units (HUs) of the corresponding pixels. Additionally, we can use pixel-spacing metadata from the input DICOM images to measure the cross-sectional area of each tissue type on the axial DICOMs. We report density and cross-sectional area of muscle, SAT, VAT, and optionally IMAT. 

\subsection{Outputs}
Our pipeline generates segmentations of bone, muscle, and adipose tissue at the T12 - L5 vertebral levels. From these segmentations, we generate several body composition measures, including Hounsfield units (HU) measures within 3D spine ROIs for T12 - L5, as well as mean HU and area within 6 axial slices corresponding to T12 - L5 centers.

Outputs from this analysis are stored within the \emph{outputs} folder. By default, this folder is placed in the top-level \OM folder. Within the outputs folder, a subfolder is created for the current run, using the date-time naming format: \emph{<Y>-<m>-<d>\_<H>-<M>-<S>}. Within this per-run folder, additional folders are created for each CT series that is processed. These folders share names with the folders that store the CT series. Within each of these per-series folders, three folders are created with the names \emph{images}, \emph{segmentations}, and \emph{metrics}. The images folder stores image outputs as PNG files with the goal of allowing for quality assurance procedures. The segmentations folder contains files that store the predicted segmentation masks. The metrics folder contains a CSV file that stores computed body composition measures, like tissue area and density.

\noindent\textbf{Images:} To enable visual inspection of the spine ROIs, we save sagittal and coronal curved planar reformations (CPRs) that pass through each of the ROIs. The path that is traversed through the 3D volumes to generate the curved planar reformations is computed using linear interpolations between the centers of the ROIs. Additionally, the pixel spacing in the superior-inferior dimension often differs from the pixel spacing in the right-left and posterior-anterior dimensions. To render the image spatially isotropic, we apply an order-3 spline interpolation.

\begin{figure}[ht]
    \centering
    \includegraphics[scale=0.5]{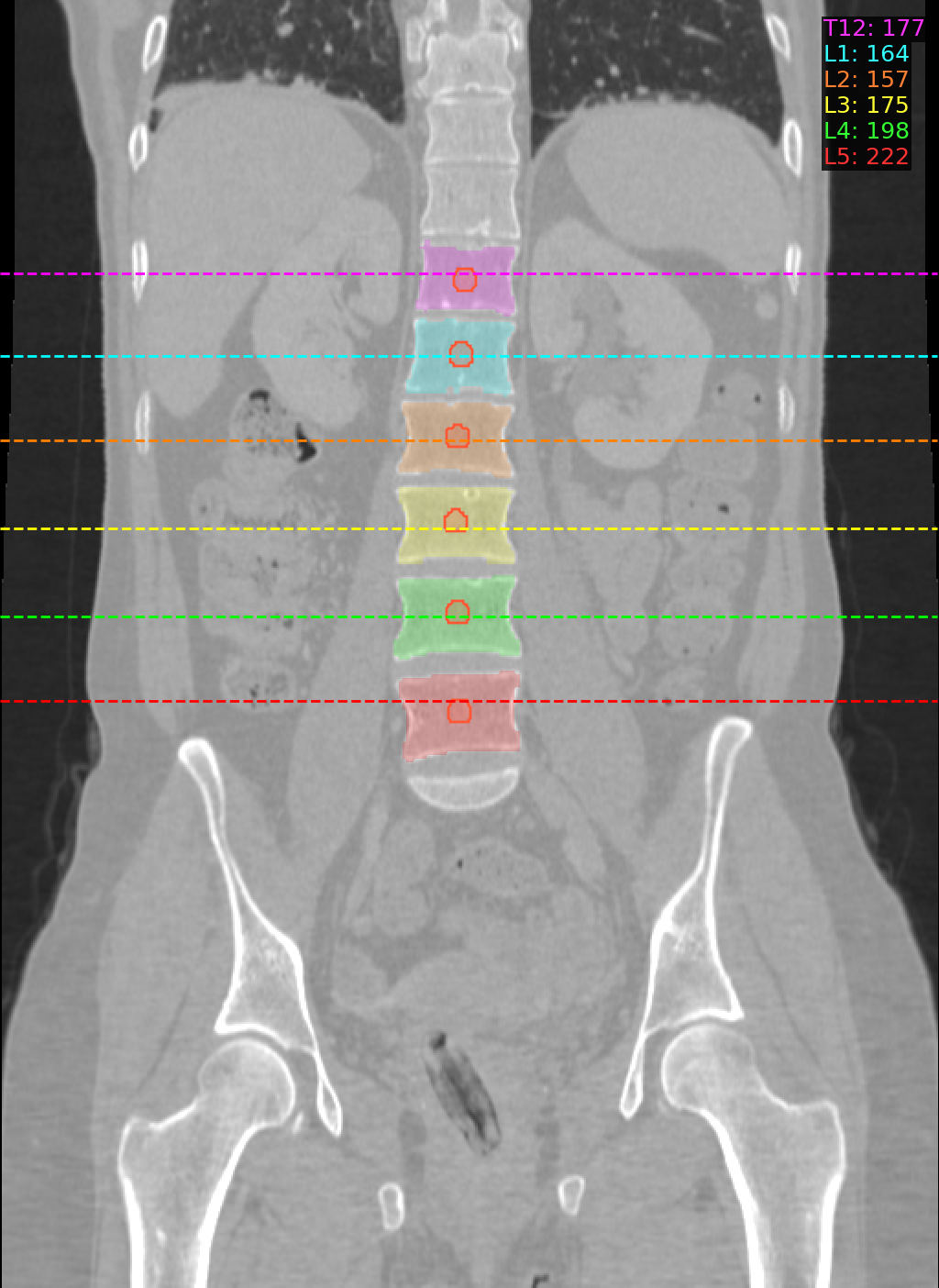}
    \caption{Coronal curved planar reformation through spherical ROIs, including the ROIs, segmentation predictions, and the median Hounsfield units within the ROIs (upper right corner).}
    \label{fig:sagittal}
\end{figure}

\begin{figure}[ht]
    \centering
    \includegraphics[scale=0.5]{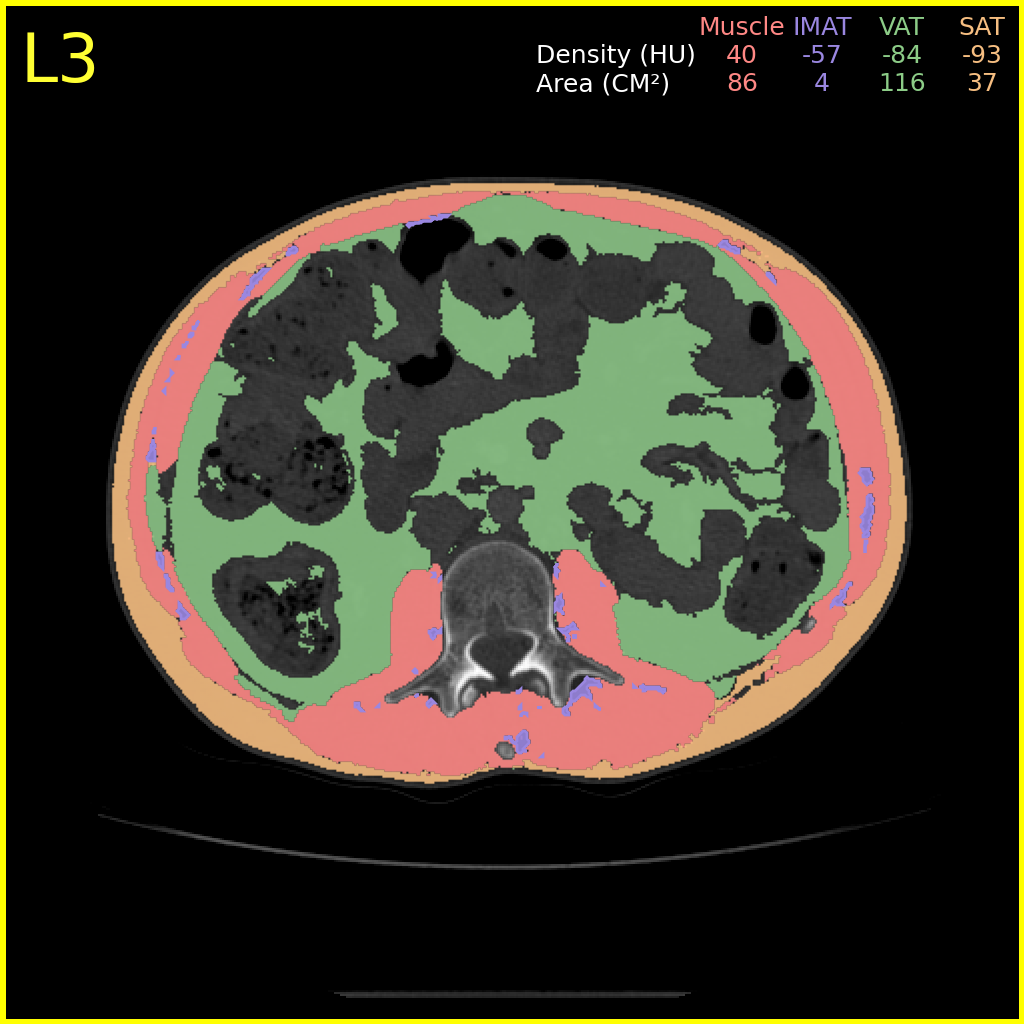}
    \caption{Muscle and adipose tissue segmentations at the L3 vertebral level, including the area (mm$^2$) and mean Hounsfield units of each tissue (upper right corner). The yellow border and yellow label (upper left corner) indicate that this axial image corresponds to the L3 vertebral level.}
    \label{fig:l3}
    
\end{figure}

Figure~\ref{fig:sagittal} shows an example coronal CPR that is stored in the \emph{images} folder. In the CPRs, we include corresponding reformations of the regions of interest (orange circles) and the segmentation predictions. The dashed lines represent the per-level superior-inferior centers. These centers are used to select axial DICOM files that are subsequently used for 2D skeletal muscle and abdominal adipose tissue segmentation. The colors of these dashed lines correspond to the border colors of the associated axial image outputs, as in the L3 axial image in Figure~\ref{fig:l3}.

Figure~\ref{fig:l3} shows an example image output of the muscle and adipose tissue segmentation model. We overlay the muscle, IMAT, VAT, and SAT segmentations in red, blue, green, and yellow respectively. These images are named using the vertebral level that they correspond to in the format \emph{<level>$\_$seg.png}. 

\begin{figure}[ht]
    \centering
    \includegraphics[scale=0.2]{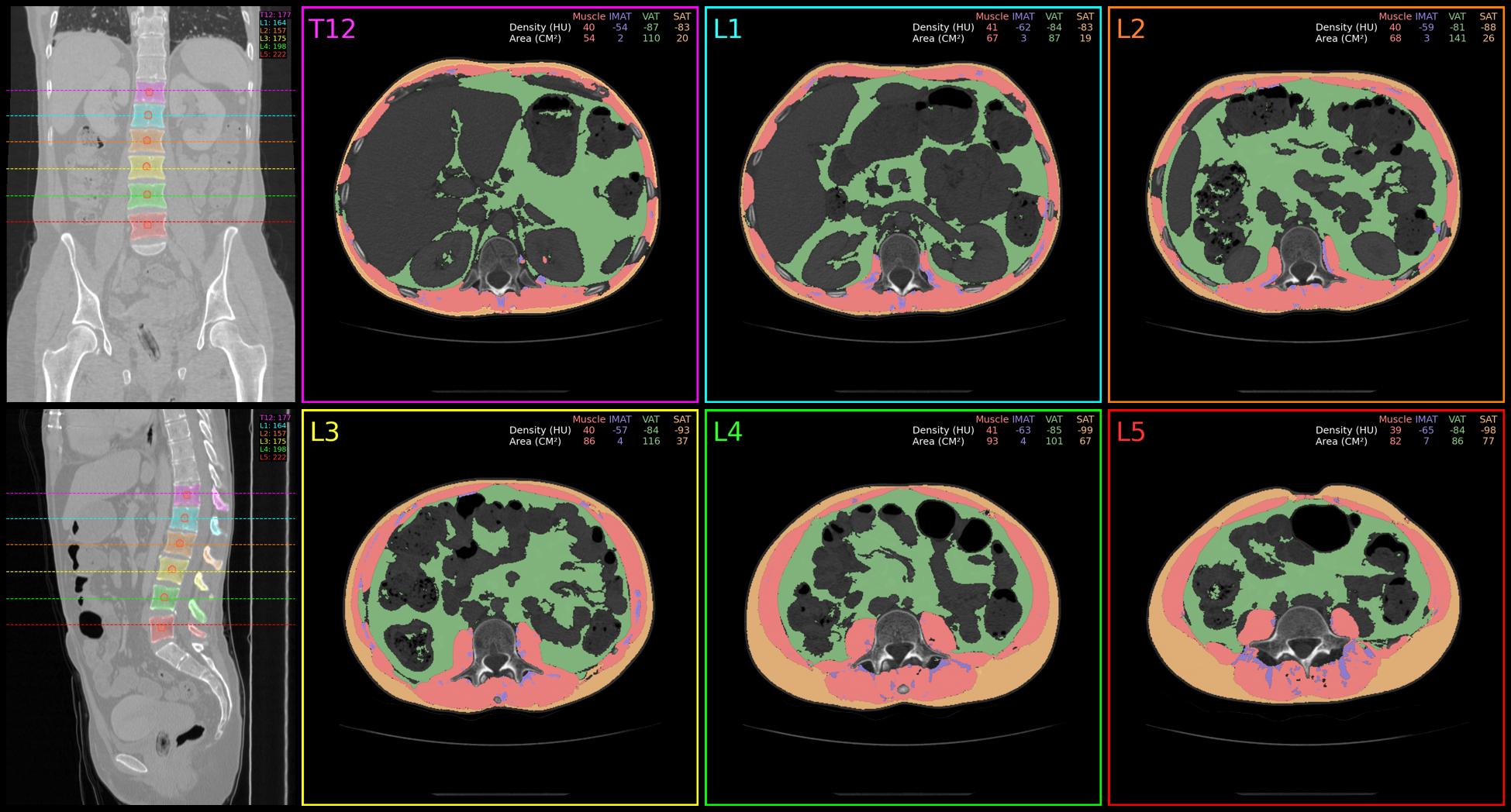}
    \caption{Visual report includes all body composition information in a PNG file.}
    \label{fig:panel}
\end{figure}

\noindent\textbf{Segmentation Files:} Within the segmentations folder, the 3D spine segmentation is saved in NIfTI format as spine.nii.gz, while the multi-level muscle and adipose tissue segmentations are saved in H5 format. 2D segmentations of muscle and adipose tissue are stored in separate H5 files for each level, named using the convention \emph{<level>$\_$seg.h5}. Within these H5 files, there is a group with the name \emph{stanford$\_$v0.0.1} that contains members corresponding to various tissues, like \emph{imat}, \emph{muscle}, \emph{sat}, and \emph{vat}. 

\noindent\textbf{CSV Metrics:} In the CSV output file, we include the paths to the DICOM files for each T12 - L5 vertebral level. For each level, we include the spine ROI Hounsfield unit measures, which are by default the median pixel within the ROIs. Additionally, for muscle, IMAT, VAT, and SAT, we include the area in cm$^2$ and the mean Hounsfield unit values.

\subsection{Basic Usage of Pipeline 1}
To run body composition analysis, use the following command. Here, the user can specify any folder on the file system and \OM will process all folders (containing only DICOM files and including > 300 files) under the specified folder. \OM will attempt to processes any such CT series that lives under the input folder and if an error is encountered, a traceback will be printed, and the pipeline will move on to processing the next series. For the initial release (v0.1.0), DICOM series are processed sequentially. To increase the level of parallelism, multiple \OM instances can be run on different compute nodes, with different input paths specified for each. 

\begin{codeshell}
bin/C2C process_3d INPUT_PATH /path/to/input/folder
\end{codeshell}

To automatically schedule a job to SLURM, you can modify the above command as follows:

\begin{codeshell}
bin/C2C-slurm process_3d INPUT_PATH /path/to/input/folder
\end{codeshell}

\section{Pipeline 2: Modular Slice-by-Slice 2D Analysis of Muscle and Adipose Tissue}
If the user wishes to bypass the spine segmentation process, we provide an interface for segmenting muscle and adipose tissue on user-specified axial DICOM slices. The same muscle and adipose tissue models that are used for \emph{Pipeline 1} are also used for \emph{Pipeline 2}. 

\subsection{Inputs}
\label{inputs2}
For \emph{Pipeline 2}, input to the 2D pipeline is a folder containing one or more DICOM files or subfolders that contain DICOM files.  

\subsection{Outputs}
The outputs of \emph{Pipeline 2} are similar to those of \emph{Pipeline 1} with some modifications. As with \emph{Pipeline 1}, the output folder is placed in the top-level \OM folder. Within this folder, we generate a per-run folder using the same date-time naming convention as is used in \emph{Pipeline 1}. Within this per-run folder, a subfolder is created with the same name as the folder used as the INPUT$\_$PATH argument. Within this folder, three folders are created with the names \emph{images}, \emph{segmentations}, and \emph{metrics}. 

\noindent\textbf{Images:} The images folder stores image outputs as PNG files, one for each DICOM file in the input. These images are named using the input DICOM name. For example, if the DICOM file is named \emph{abcd.dcm}, the corresponding image file would be named \emph{abcd.png}.

\noindent\textbf{Segmentation Files:} The segmentations folder contains H5 files that store the predicted segmentation masks. As with the image files, there is one H5 file per DICOM input file and these files are named using the input DICOM name. 

\noindent\textbf{CSV Metrics:} The computed body composition metrics are stored in a CSV file in the metrics folder. For each DICOM file and each tissue, area (cm$^2$) and mean Hounsfield units are reported. 

\subsection{Basic Usage of Pipeline 2}
To run \emph{Pipeline 2}, use the following command where the path points to the folder described in \ref{inputs2}.

\begin{codeshell}
bin/C2C process_2d INPUT_PATH /path/to/input/folder
\end{codeshell}

As with \emph{Pipeline 1}, if you are using a SLURM environment, you can replace the above command with this command to automatically submit a job to SLURM:

\begin{codeshell}
bin/C2C-slurm process_2d INPUT_PATH /path/to/input/folder
\end{codeshell}

\section{Preliminary Validation}

\subsection{Spine}
We compared the superior-inferior (vertical) centers generated by the TotalSegmentator spine segmentation model~\cite{https://doi.org/10.48550/arxiv.2208.05868} to vertical centers computed using 40 Stanford tertiary center emergency department intravenous contrast-enhanced test set cases. This externally validates the TotalSegmentator model, as well as our methods for extracting vertical centers and ROIs. Our labels include only the vertebral bodies, while TotalSegmentator includes the spinous/transverse processes. Nonetheless, comparing vertical centers at each level gives the mean errors in column 1 of Table \ref{table:validation_spine}. Moreover, we compared our method for extracting mean HUs using our spine ROIs (10x10x10 pixel cubic) to the mean HUs generated from equivalently shaped ROIs positioned at the centers of our labeled vertebral bodies. We achieved mean errors in column 2 of Table \ref{table:validation_spine}. Corresponding relative errors are shown in column 3 of Table \ref{table:validation_spine}. This validates our method for isolating the vertebral bodies and positioning the ROIs as described in Figure~\ref{fig:spine_rois}.

\begin{table}[ht]
\centering
\caption{External validation of the Total Segmentator spine model and ROIs on 40 cases. Values presented as mean (median). The errors here are mean (median) absolute errors.}
\begin{tabular}{c c c c} 
 \hline & \\[-1.5ex]
 Level & Vertical Center Error (mm) & ROI HU Error  & ROI HU \% Error \\
 \hline & \\[-1.5ex]
 T12 & 4.93 (1.82) & 4.63 (1.62) & 2.24 (0.82)\\ 
 L1 & 5.35 (2.39) & 2.94 (1.29) & 1.31 (0.55)\\
 L2 & 4.78 (2.24) & 3.28 (1.52) & 1.61 (0.93)\\
 L3 & 3.38 (2.00) & 3.43 (1.93) & 1.60 (0.86) \\
 L4 & 2.68 (0.82) & 7.10 (1.79) & 2.57 (1.04)\\
 L5 & 4.30 (2.95) & 3.66 (1.74) & 1.61 (0.97)\\
 \hline
\end{tabular}
\label{table:validation_spine}
\end{table}

\subsection{Muscle and Adipose Tissue}

At the L3 vertebral level, our \verb|stanford_v0.0.1| muscle and adipose tissue model achieved mean (standard deviation) Dice scores of 0.97 (0.03), 0.96 (0.05), and 0.97 (0.02) for muscle, VAT, and SAT respectively on a held-out internal test set of 40 intravenous contrast enhanced CT scans from our tertiary center emergency department. Furthermore, the error in computing tissue Hounsfield units and cross-sectional area averaged below 1\% and 2\%, respectively, for all three segmented tissues~\cite{chaves2021opportunistic}. Table \ref{table:validation} lists the results of external validation for muscle, VAT, and SAT on 20 CTs from another institution.

\begin{table}[ht]
\centering
\caption{External validation of the stanford\_v0.0.1 muscle and adipose tissue model on 20 external cases. Values presented as Dice mean (standard deviation). For reference, the following values were achieved on the internal test set at the L3 vertebral level: 0.97 (0.03), 0.96 (0.05), and 0.97 (0.02) for muscle, VAT, and SAT respectively.}
\begin{tabular}{c c c c} 
\hline & \\[-1.5ex]
Level & Muscle & VAT & SAT \\
\hline & \\[-1.5ex]
 T12 & 78.0 (9.6) & 90.9 (8.0) & 88.8 (22.1)\\ 
 L1 & 85.5 (10.0) & 92.7 (7.2) & 89.3 (22.4)\\
 L2 & 92.1 (7.8) & 94.0 (6.4) & 90.5 (21.7)\\
 L3 & 94.7 (5.9) & 94.6 (6.7) & 93.2 (14.8) \\
 L4 & 94.4 (4.9) & 93.6 (8.0) & 94.8 (9.7)\\
 L5 & 84.7 (10.3) & 93.6 (6.8) & 95.0 (9.9)\\
 \hline
\end{tabular}
\label{table:validation}
\end{table}

\section{Conclusion}
We have designed \OM to make it user-friendly, efficient, and comprehensive. As such, we hope that \OM will contribute to an increased ubiquity of body composition analysis in large scale research studies and clinical settings. We are committed to regularly updating the tool with new features to keep it up-to-date with the latest advancements. We plan to add health outcome risk scores, as well as radiomic analyses. We encourage the community to submit pull requests and issues to help us improve the tool and increase its value to the community.

\bibliographystyle{abbrv}
\bibliography{refs} 

\end{document}